\documentclass[final,twocolumn]{svjour3}       
\smartqed  

\usepackage{url}
\usepackage{graphics}
\usepackage[pdftex]{graphicx}
\usepackage{epstopdf}
\usepackage{algorithmic}
\usepackage{algorithm}
\graphicspath{{figures/}}

\DeclareGraphicsExtensions{.jpg,.pdf,.tif,.png,.tiff,.eps}

\RequirePackage{fix-cm}

\usepackage[numbers,sort&compress]{natbib}
\usepackage{lineno}
\usepackage{algorithmic}
\usepackage[cmex10]{amsmath}
\usepackage{amssymb}
\usepackage{mdwmath}
\usepackage{mdwtab}
\usepackage{dblfloatfix}

\newcommand{\ccc}{\mathbf{c}}
\newcommand{\sss}{\mathbf{s}}
\newcommand{\sssp}{\mathbf{s}^{\prime}}
\newcommand{\zzz}{\mathbf{z}}
\newcommand{\zzzp}{\mathbf{z}^{\prime}}
\newcommand{\uuu}{\mathbf{u}}
\newcommand{\uuup}{\mathbf{u}^{\prime}}

\usepackage{appendix}
\DeclareMathOperator{\sgn}{sgn}
\usepackage{setspace}
\usepackage{baskervald}
\begin{document}

\title{Iterative graph cuts for image segmentation with a nonlinear statistical shape prior}
\titlerunning{Fast shape regularized segmentation with MM-graphcuts}
\journalname{Journal of Mathematical Imaging and Computer Vision}
\date{\today}
\author{Joshua C. Chang \and Tom Chou }
\institute{J.C. Chang \at
  Mathematical Biosciences Institute \\
The Ohio State University\\
Jennings Hall, 3rd Floor\\
1735 Neil Avenue\\
Columbus, Ohio 43210\\
  \email{chang.1166@mbi.osu.edu} \\
  \and
  T. Chou \at
  UCLA Biomathematics and Mathematics \\
BOX 951766, Room 5303 Life Sciences \\
Los Angeles, CA 90095-1766 \\
  \email{tomchou@ucla.edu}
 }

\maketitle

\begin{abstract}
Shape-based regularization has proven to be a useful method for delineating objects within noisy images where one has prior knowledge of the shape of the targeted object. When a collection of possible shapes is available, the specification of a shape prior using kernel density estimation is a natural technique. Unfortunately, energy functionals arising from kernel density estimation are of a form that makes them impossible to directly minimize using efficient optimization algorithms such as graph cuts.  Our main contribution is to show how one may recast the energy functional into a form that is minimizable iteratively and efficiently using graph cuts.

\end{abstract}
\section{Introduction}
Graph cuts provides an ingenious technique for image segmentation that relies on transforming the problem of energy minimization into the problem of determining the maximum flow or minimum cut on an edge-weighted graph. By using graph cuts, segmentations can be found efficiently in low-order polynomial time. When images are noisy, segmentation performed using only image data produces poor results. In these cases, regularization is needed. 

Typically, when one is analyzing images, one has expectations for what he or she is looking for -- people, birds, cars, cows, bacteria, whatever the object may be. In these cases it is natural to try to restrict segmentation to results that match the shape of the desired object.  Within any class of objects there is often variability in shape. Such variability can arise even when dealing with shape-invariant objects, for example by changes in pose or perspective. Probabilistic expression of shape knowledge  is a natural way of capturing this variability.

\subsection{Related prior work}

In~\citet{chang2012tracking}, we presented a technique for tracking an object whose boundary motion is modeled using the level-set method. Such objects do not retain any particular a-priori shape. We remarked that one could adapt our regularization method to the segmentation of objects that do retain their shapes. The purpose of this article is to explicitly present a method for performing such a task.

Our method is grounded in the theory of graph cuts-based image segmentation with shape-based regularization, where segmentation is performed using a-priori shape knowledge. The prior literature on this subject is vast. Briefly, we mention a few influential articles. \citet{slabaugh2005graph}  developed a graphcuts-based method for segmentation assuming that the desired object is an ellipse.  Allowing for more flexibility, \citet{freedman2005interactive} provided a method to segment any particular shape based on the signed distance embedding. Some methods have concentrated on looking at shapes with certain characteristics such as compactedness or local convexity~\cite{veksler2008star,das2009semiautomatic}.  Other methods have treated the task of space specification as a statistical problem.

Statistical shape priors have taken many forms in the literature~\cite{malcolm2007graph,dambreville2008framework,zhu2007graph}. Two of the most common forms of shape representation are kernel-PCA and kernel-based level-set embeddings. Both methods learn a probability distribution for the target shape based on a set of training shapes or templates. They differ in how they use the training images. Kernel PCA-based methods project the training templates into principal-component space and define probability distributions over the resulting vector spaces. Level-set-based methods define a distance between implicitly embedded shapes and define a probability distribution with respect to the distance.

One major disadvantage of the PCA-based approach is that the resulting probability distribution for the shape is Gaussian. Given a set of training templates, this approach will tend to favor shapes that are more similar to the space-averaged shape, thereby potentially making certain training shapes improbable. This problem also exists for exponential-family shape priors where the log-prior is a linear combination of the training templates. Combinations of shapes may not be valid shapes~\cite{cremers2008shape}, and under this approach certain training templates may become improbable as well. Instead, we opt for a kernel-density based approach. 

Kernel density estimation (KDE) is a non-parametric method for the estimation of probability distributions. Its main advantage is its flexibility and its ability to capture multiple modes. Our method here, and that given in \citet{chang2012tracking}, can be described as a graph-cuts adaptation of the method of~\citet{cremers2006kernel}. In their paper, they treated shape-prior specification as a density estimation problem and used KDE to derive a prior from an ensemble of training shapes. The resulting energy functionals, being nonlinear, require some intervention to use in a graph-cuts framework. 
  
In this manuscript, we make the main contribution of demonstrating the use of majorization-minimization (MM) for the iterative relaxation of a nonlinear energy functional using graph-cuts. The MM algorithm is a generalization of the well-known expectation-maximization (EM) algorithm. It is often useful for linearizing  nonlinear objective functions. We use MM to find a surrogate minimization problem that is solved using graph-cuts. In doing so, we expose a mathematical link between the
energy functional of~\citet{cremers2006kernel}, and the linear energy functionals commonly used for graph-cuts with statistical priors~\cite{malcolm2007graph,dambreville2008framework,zhu2007graph}.
 The subsequent use of graph cuts results in a computationally more efficient approach relative to that of using level set approaches~\cite{grady2009piecewise}. 
 
\section{Mathematical Method}

\subsection{Shapes}
Signed-distance functions provide a handy tool to represent shapes mathematically. Shapes or regions 
$\Omega\subset\mathbb{R}^d$ can be implicitly represented by a
 function $\phi_{\Omega}(\sss):\mathbb{R}^d\rightarrow\mathbb{R}$,
which for every $\sss\in\mathbb{R}^d$ 
 is the \emph{signed Euclidean distance} from $s$ to the boundary of $\Omega$.
In this paper we take the convention that $\phi_{\Omega}(\sss)$ is positive if
$\sss\in\Omega$, and negative if $\sss\in\mathbb{R}^d\setminus\Omega$.

\begin{figure}
\begin{center}
\includegraphics[width=21pc]{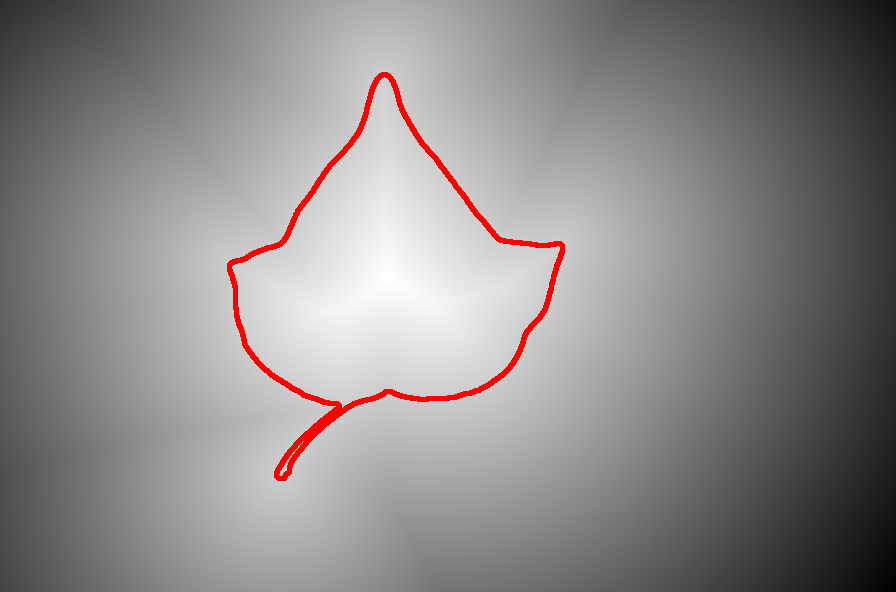}
\caption{{\bf Embedding of shapes using signed Euclidean distance functions.} A shape (pictured is a leaf) is embedded as a signed Euclidean distance function where values inside are the positive distance from the boundary, and values outside are negative distances from a boundary. The boundary of the leaf is the $0$-level set of the function (red outline).}
\end{center}
\end{figure}

In many applications, since the shape of an object is only approximately known, it 
is advantageous to represent the knowledge of the shape probabilistically. 
We use a kernel density estimate of
 the distribution of possible shapes embedded as a collection of individual discretized level sets.
   Let us first denote 
$\chi_\Omega$ the characteristic function for a region $\Omega$,
\begin{equation}
\chi_\Omega(\sss)=\begin{cases}
1, & \sss\in\Omega; \\
0, & \sss\not\in\Omega.
\end{cases}
\end{equation}
Then, for  two shapes $\Omega$ and $\Lambda$,
embedded as signed distance functions $\phi_{\Omega}$ and $\phi_{\Lambda}$,
 we introduce an (asymmetric) pairwise shape energy 
 \begin{align}\label{eqn:shapedistance}
U_{\textrm{shape}}(\Omega,\Lambda)&=\overbrace{\int_{\mathbb{R}^d} \left|\chi_\Omega(\sss)-\chi_\Lambda(\sssp)\right|\left|\phi_\Lambda (\sssp)\right|^\lambda d\sss}^{\textrm{mass mismatch}}\nonumber\\
&\qquad + \overbrace{\oint_{\partial\Omega} \left|\phi_{\Lambda}(\zzzp)\right|^\lambda d\zzz}^{\textrm{boundary mismatch}},
\end{align}
where $\lambda\geq 0$.
This expression is a generalization of other shape energies seen in the literature, where  commonly $\lambda=0,1,2$~\cite{malcolm2007multi,dambreville2008framework,zhu2007graph,cremers2008shape,vu2008shape,lempitsky2008image,heimann2009statistical,lempitsky2012branch}. 
Although symmetric shape energies are desirable~\cite{cremers2003pseudo}, for general use in our optimization scheme we require an asymmetric energy that does not depend on the Euclidean-distance shape embedding of the evolving segmentation. If one uses $\lambda=0$, then it is possible to make this energy symmetric with small modifications (see Discussion).

 To make this energy robust to changes in scale, orientation, and location, a transformation $\sssp=\mathbb{T}(\sss)$ that maps points $\sss$ in the reference frame of $\Omega$ to points $\sssp$ in the reference frame of $\Lambda$ is needed. In Appendix~\ref{sec:newton}, we provide a method for finding rigid transformations of the form \[ \mathbb{T}(\sss)=\alpha \mathbf{R}(\boldsymbol\omega) \left( \sss - \boldsymbol\ccc\right)\]
 where $\alpha$ is a scaling factor, $\mathbf{R}(\boldsymbol\omega)$ is a rotation matrix, and $\boldsymbol\ccc$ is a translation vector. These three parameters are chosen to minimize Eq.~\ref{eqn:shapedistance} (See Appendix~\ref{sec:newton} for details on calculating these parameters). This method worked well for our examples; however, other shape alignment schemes may be used~\cite{lee1994nonlinear,jiang2008robust,viola1997alignment,tabbone2003binary,belongie2002shape}.

After transformation, we can use the transformation-minimized energy in Eq.~\ref{eqn:shapedistance} to 
assemble references shapes  $\{\Omega_j\}_{j=1}^J$ into a probability distribution by using a kernel of the form
  \begin{equation}\label{eqn:kernel}
 K(\Omega,\Lambda)= \sqrt{\frac{\beta}{2\pi}}e^{-\beta{U_{\textrm{shape}}(\Omega,\Lambda)}}.
 \end{equation}
We can then represent a probability distribution over shapes, $p(\Omega)$ according to:
\begin{equation}
p(\Omega)\propto \sum_{j=1}^J w_j K(\Omega,\Omega_j),
\label{eqn:prior}
\end{equation}
where  $w_j$ are weighting coefficients that sum to one.
 This representation of the prior is a kernel density estimate~\cite{silverman1986density} (KDE)
 of the distribution of shapes. We  set $\beta$, the reciprocal temperature,  to be the scale-normalized
kernel width~\cite{silverman1986density,cremers2006statistical}:
\begin{equation}
\beta = \left[\sum_{j=1}^J \frac{w_j}{s_j^\lambda} \min_{k\neq j} {U_{\textrm{shape}}(\Omega_k,\Omega_j)}\right]^{-1},
\end{equation}
where $s_j$ is the inertial scale of a shape $\Omega_j$ as defined in \citet{jiang2008robust}. Note that this expression of the shape prior is potentially multi-modal.

\subsection{Generative image modeling}

The task is to identify an object $\Omega$ in an image $I:S\subset\mathbb{R}^d\to\mathbb{R}^q$, where $q$ refers to the dimension of the color space used.
We model an image probabilistically
with distributions of intensity values conditional on a parameter vector $\boldsymbol\theta$, and region:
\begin{equation}
p(I(\sss)|\boldsymbol\theta)=\begin{cases}
p_\Omega(I |\boldsymbol\theta) & \sss\in\Omega \quad\text{(foreground)}; \\
p_\Delta(I |\boldsymbol\theta) & \sss\in\Delta = S\setminus\Omega \quad\text{(background)}.
\end{cases}
\end{equation}
Information about $\boldsymbol\theta$ and
$\Omega$ can be incorporated as prior probability distributions $p(\boldsymbol\theta|\Omega)$ and $p(\Omega)$. 
With these prior distributions, we can write the joint posterior distribution $p(\Omega,\boldsymbol\theta|I)\propto
p(I|\Omega,{\boldsymbol\theta})p(\boldsymbol\theta|\Omega)p(\Omega)=e^{-U(\Omega,\boldsymbol\theta)}$,
where $U(\Omega,\boldsymbol\theta)$ is an energy (not to be confused with $U_{\textrm{shape}}$).

 We wish to infer the
segmentation $\Omega$ by maximizing the posterior probability relative
to $\Omega$. To this end, we will maximize the logarithm of the
posterior, or equivalently, minimize the energy
\begin{align}
\lefteqn{U(\Omega,\boldsymbol\theta)=-\log[ \overbrace{ p(I|\Omega,{\boldsymbol\theta})}^{\text{likelihood}}\overbrace{p(\boldsymbol\theta|\Omega)p(\Omega)}^{\text{prior}}] }\nonumber\\
&=-\sum_{\sss\in\Omega}\log p_\Omega(I(\sss)|\boldsymbol\theta)-\sum_{\sss\in\Delta}\log p_\Delta(I(\sss)|\boldsymbol\theta) \nonumber\\
&\quad  -\underbrace{\log\sum_{j=1}^J w_j K(\Omega,\Omega_j)}_{\textrm{Eqs.}~\ref{eqn:kernel},~\ref{eqn:prior}} \nonumber\\
& \qquad- \log p(\boldsymbol\theta|\Omega) + \text{const}.
\label{eqn:energysum}
\end{align}

\subsection{Majorization-Minimization}\label{sec:major}

The energy functional in Eq.~\ref{eqn:energysum} resembles the energy functional given by 
\citet{cremers2006kernel}, where level-set based gradient descent is used for relaxation. Unfortunately, optimization performed in this manner is slow, since only small deformations of an evolving contour occur in each iteration.

 We take an iterative two-step approach to minimizing this energy. Given the segmentation estimate $\Omega^{(n)}$ in the n-th step, we minimize the energy with respect to $\boldsymbol\theta$~\cite{o2004bayesian}.

Given $\boldsymbol\theta^{(n)}$, we find the optimal $\Omega^{(n+1)}$ by iteratively minimizing 
\begin{align}
\lefteqn{Q(\Omega|\Omega^{(n)})=-\sum_{\sss\in\Omega}\log p_\Omega(I(\sss)|\boldsymbol\theta)-\sum_{\sss\in\Delta}\log p_\Delta(I(\sss)|\boldsymbol\theta)} \nonumber\\
 &\quad+ \underbrace{\frac{\beta}{2} \sum _{j=1}^J c_j^{(n)}U_{\textrm{shape}}(\Omega,\Omega_j)}_{\textrm{majorization}},
\label{eqn:mmenergy}
\end{align}
where
\begin{align}
c_j^{(n)}= \frac{w_j K(\Omega^{(n)},\Omega_j)}{\sum_{k=1}^J w_kK(\Omega^{(n)},\Omega_k)}.
\label{eqn:mmweight}
\end{align}
 In Eq.~\ref{eqn:mmenergy}, we have replaced logarithm-sum term of~Eq. \ref{eqn:energysum} 
with a linearized majorization term, by exploiting the concavity of the logarithm function.
The majorization term defines a surrogate minimization
problem that is more easily solved -- in our case using graph cuts -- albeit at the price of iteration.
Such an algorithm is commonly known as a majorization-minimization or
MM algorithm~\cite{hunter2004tutorial} (see Appendix~\ref{sec:graphembed}). 
This approach is generally possible for other energies that contain components of the form
$f( J[\Omega]), $
where the function $f:\mathbb{R}\to\mathbb{R}$ is convex, and $J$ is a functional 
that is dependent on $\Omega$.

\subsection{Graph cuts for surrogate energy relaxation}\label{sec:gc}

Here, we describe minimization of the surrogate energy in
Eq.~\ref{eqn:mmenergy} using graph cuts. Graph cut methods have
their grounding in combinatorial optimization theory, and are
concerned with finding the minimum-cost cut in an undirected graph. A cut
is a partition of a connected graph into two disconnected sets. The
cost of a cut is the sum of the edge weights along a cut, and a
max-flow min-cut algorithm finds the cut with the lowest cost. To use
graph cuts for image segmentation, we must express our energy function
in terms of edge-weights on a graph.
Following \citet{boykov2001interactive}, 
 we begin by expressing the energy given in
Eq.~\ref{eqn:mmenergy} as a function of the vertices $V$ and edges
$\mathcal{E}$ of a graph $\mathcal{G}=(V,\mathcal{E})$:
\begin{equation*}
U(\mathcal{G})=\sum_{\sss\in V} U_V(\sss)+\sum_{(\sss,\uuu)\in\mathcal{E}} U_{\mathcal{E}}(\sss,\uuu).
\end{equation*}
 
\begin{figure*}
\begin{center}
\includegraphics[width=0.85\textwidth]{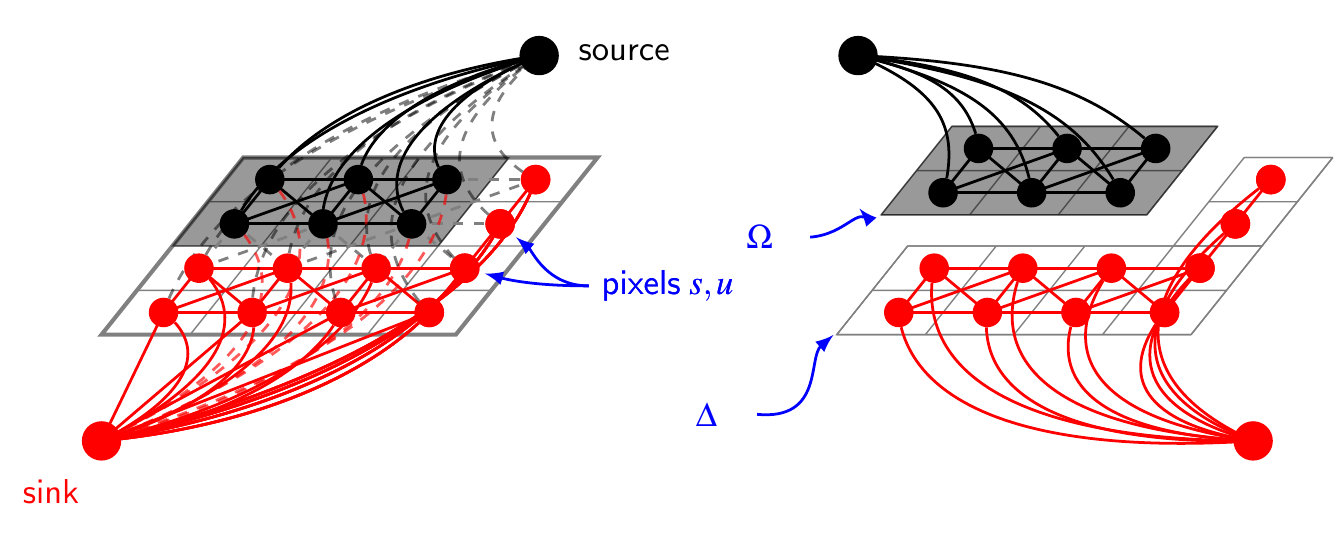}
\caption{ \textbf{\emph{(left)} Embedding of segmentation into a graph} In the graph-cuts framework, the problem of segmentation is transformed into the problem of finding the minimum cut in an edge-weighted graph. The energy to minimize is rewritten as a sum of atomic terms and a sum of neighbor-interaction terms. The edges are weighted such that the cut is the energy of the corresponding segmentation. \textbf{\emph{(right)} Resulting segmentation.} The cutting of the graph yields two disconnected node sets where $\Omega$ is the foreground and $\Delta$ is the background.}\label{fig:gc}
\end{center}
\end{figure*}

   In the graph that we construct,
each pixel is assigned a node, and edges exist
between nodes representing neighboring pixels (Fig~\ref{fig:gc}). The
neighbor edges are known as neighbor-links (n-links).
 Two special nodes called the source (foreground) and sink (background) are added,
along with edges connecting these nodes to each pixel node (Fig~\ref{fig:gc}). 
These edges are known as terminal-links (t-links).

 We want to infer 
an unknown two-coloring on the nodes of 
the graph that represents inclusion of a node
$s$ into either the foreground set $\Omega$, or the background set $\Delta$. This inference involves splitting the graph into two parts (Fig~\ref{fig:gc}), where a pixel's connection to the source represents inclusion into the foreground set. 

Discretizing the shape divergence~(Eq.~\ref{eqn:shapedistance}) 
 over an eight-connected neighbor graph yields
 \begin{align}\label{eqn:discreteshapedistance}
\lefteqn{U_{\textrm{shape}}(\Omega,\Lambda)=}\nonumber\\ & \sum_\sss \overbrace{\left|\chi_{\Omega}(\sss)-\chi_{\Lambda}(\sssp)\right|}^{\text{indicator of label mismatch}} \left|\phi_{\Lambda}\left(\sssp \right)\right|^\lambda  \nonumber\\
&+ \sum_{(\sss,\uuu)\in\mathbf{N}}\overbrace{\left|(\chi_{\Omega}(\sss)(1-\chi_{\Omega}(\uuu))+\chi_{\Omega}(\uuu)(1-\chi_{\Omega}(\sss))\right|}^{\textrm{ indicator of } \sss,\uuu \textrm{ lying across boundary of }\Omega}\nonumber\\
&\qquad\qquad \times \frac{\pi}{8||\sss-\uuu ||}\left|\phi_{\Lambda}\left(\frac{\sssp+\uuup}{2} \right)\right|^\lambda ,
\end{align}
where $\mathbf{N}$ refers to the set of tuples of nodes that are neighboring in space. The level set function $\phi_\Lambda^\lambda$ is taken to be the Euclidean distance embedding of $\Lambda$, transformed to the center, scale, and orientation of $\Omega$. As mentioned before, we use an asymmetric representation
of the shape divergence since the signed-distance function is a property of a global segmentation and is not obtainable from local labeling. 

Using this expression, one sees that
\begin{align}
U_V(\sss) &=-\chi_{\Omega}(\sss)\log p_\Omega(I(\sss)|\boldsymbol\theta)\nonumber\\
&\quad-\left(1-\chi_{\Omega}(\sss)\right)\log p_\Delta(I(\sss)|\boldsymbol\theta) \nonumber\\
&\quad + \frac{\beta}{2}\sum_{j=1}^J c_j^{(n)}\left|\phi_{\Omega_j}\left(\sssp \right)\right|^\lambda \left|\chi_{\Omega}(\sss)-\chi_{\Omega_j}(\sssp)\right|
\end{align}
and
\begin{align}
\lefteqn{U_{\mathcal{E}}(\sss,\uuu)=}\nonumber\\ &
\quad\frac{\pi\beta}{16||\sss-\uuu||}\left[ \vphantom{\sum_{j=1}^J}\sum_{j=1}^J c_j^{(n)}\left|\phi_{\Omega_j}\left(\frac{\sssp+\uuup}{2}
  \right)\right|^\lambda\right.\nonumber\\ 
&\quad\left.\times\Big|\chi_{\Omega}(\sss)\left(1-\chi_{\Omega}(\uuu)\right)+\chi_{\scriptsize\Omega}(\uuu)\left(1-\chi_{\Omega}(\sss)\right) \Big|
 \vphantom{\sum_{j=1}^J}\right].
\label{eqn:energyedge}
\end{align}
The neighbor energy given in Eq.~\ref{eqn:energyedge} obeys the \emph{submodularity property},
that is, 
\begin{align*}
\lefteqn{U_\mathcal{E}\left(\sss\in\Omega,\uuu\not\in\Omega \right) + U_\mathcal{E}\left(\sss\not\in\Omega,\uuu\in\Omega \right) }\nonumber\\
\quad&= \frac{\pi\beta}{8||\sss-\uuu||}\sum_{j=1}^J c_j^{(n)}\left|\phi_{\Omega_j}\left(\frac{\sssp+\uuup}{2} \right) \right|^\lambda\\
\quad&\geq U_\mathcal{E}\left(\sss\in\Omega,\uuu\in\Omega \right) + U_\mathcal{E}\left(\sss\not\in\Omega,\uuu\not\in\Omega \right) \\
\quad&=0.
\end{align*}
As a result, the energy (Eq.~\ref{eqn:mmenergy}) is minimizable with
graph-cuts in polynomial time~\citep{boykov2004experimental, freedman2005energy}.

To embed $U_V(\sss)$ into the graph, we set the weights of the t-links between each pixel and the source to the following
\begin{align}
\lefteqn{w(\sss,\textrm{source})= -\log p_\Delta(I(\sss)|\boldsymbol\theta)} \nonumber  \\
&\quad +\frac{\beta}{2}\sum_{j=1}^J c_j^{(n)}\chi_{\Omega_j}(\sssp)\left|\phi_{\Omega_j} \left(\sssp \right) \right|^\lambda,
\end{align}
and the weights of the t-links between each pixel and the sink to the following
\begin{align}
\lefteqn{w(\sss,\textrm{sink})= -\log p_\Omega(I(\sss)|\boldsymbol\theta)} \nonumber\\
&\qquad + \frac{\beta}{2}\sum_{j=1}^Jc_j^{(n)}(1-\chi_{\Omega_j}(\sssp))\left|\phi_{\Omega_j}\left(\sssp \right) \right|^\lambda.
\end{align}

The cutting of the edge from an $\sss$ to the source implies that
$\sss\in\Delta$, so it adds to the cost of the cut by the contribution
of $\sss$ into the total energy as if $\sss\in\Delta$. In other words, these weights
can be interpreted as a pixel's strength of belonging to each region. 

To embed $U_\mathcal{E}$ into the graph, we set the n-links between pairwise neighboring pixels $\sss$ and $\uuu$ to
\begin{equation}
 \frac{\pi\beta}{16||\sss-\uuu||}\sum_{j=1}^J c_j^{(n)}\left|\phi_{\Omega_j}\left(\frac{\sssp+\uuup}{2} \right) \right|^\lambda. 
\end{equation}
Our surrogate energy is now minimized by finding the minimum cut of
the graph. For details on how to perform this optimization, we refer
the reader to~\citet{boykov2004experimental}. 

 \begin{algorithm}
 \caption{MM-Graphcut algorithm}
\label{alg:mmgc}
 \begin{algorithmic}
 \STATE Obtain initial guess of image intensity model parameters $\boldsymbol\theta^{(0)}$.
 \STATE Obtain initial guess of the segmentation $\Omega^{(0)}$, by segmentation without a shape prior.
 \STATE Align and rescale the shape prior templates to $\Omega^{(0)}$.
 \STATE $n=1$
 \WHILE{$U_{\textrm{shape}}(\Omega^{(n)},\Omega^{(n-1)})>\texttt{tol}$}
 \STATE Reweight the edges and perform max flow to estimate $\Omega^{(n+1)}$
 \STATE Re-estimate the image intensity parameters $\boldsymbol\theta^{(n+1)}$ (as needed)
 \STATE Re-align the shape prior templates to the guess $\Omega^{(n)}$ (as needed)
 \STATE $n++$
 \ENDWHILE
 \end{algorithmic}
 \end{algorithm}
 
\section{Results}

\begin{figure*}
\begin{center}
\includegraphics[width=\textwidth]{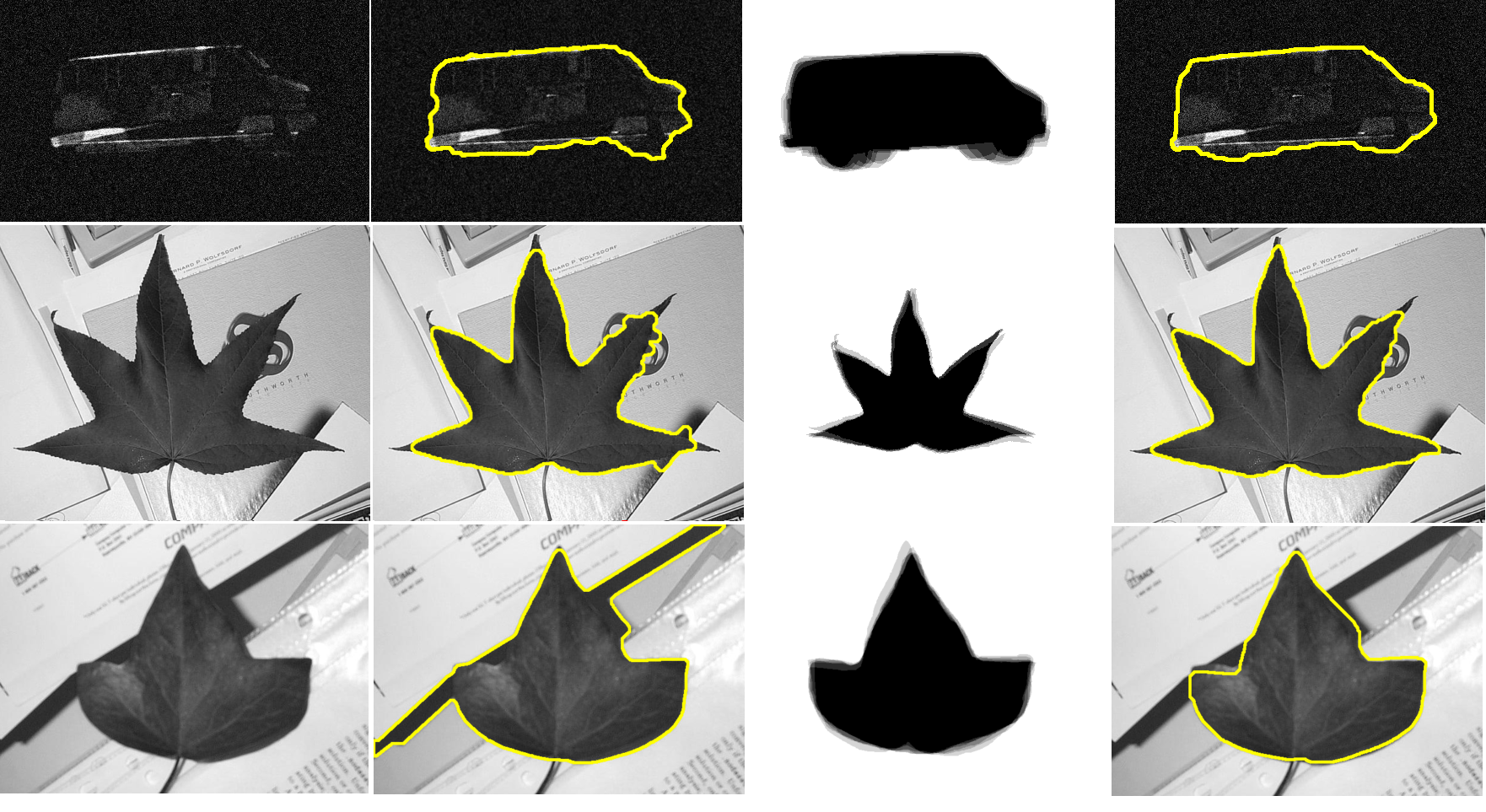} 
\caption{\textbf{Shape-regularized segmentation} \emph{From left to right: original image, length-penalized segmentation, formation of a statistical shape prior using a collection of templates, shape-regularized segmentation.} Segmentation of the original image of a van without shape information misses the wheels. A collection of templates taken together define a probabilistic shape prior. Such representation of shapes is useful since it can account for intrinsic variation in a class of shapes, as well as variations that result from changing pose. In these templates, silhouettes of a van are taken in different poses.}
\label{fig:fig3}
\end{center}
\end{figure*}

We have implemented our method in Java using the Fiji application programming interface~\cite{schindelin2008fiji} and
tested it on the regularization of luminosity-based segmentation of a variety of objects in images (Fig~\ref{fig:fig3}).
For comparison, we have obtained results of segmentation using a shape-free global length penalty~\citep{el2007graph}.  
In these examples, we have used Laplacian statistics for $p(I|\boldsymbol\theta)$ and uninformative priors for $\boldsymbol\theta$ (by setting $p(\boldsymbol\theta)=1)$. The parameter $\lambda$ was set to $2$, and the inverse-temperature $\beta$ was set to $\tau^2$. The algorithm was terminated when the labeling became stationary.

The first images we processed were low-contrast noisy images of a van. A van (Dodge B2500) was driven across a bumpy road and videotaped as it entered and exited the camcorder's field of view. Snapshots of the recording were randomly extracted, showing the van in different regions an of the image, under different orientations and different scale factors. We manually constructed a shape prior for the van by hand-drawing five templates of a van in many poses. These templates of the van were aligned using the method detailed in Appendix~\ref{sec:newton}. Shape-free segmentation of the van captured the dorsal aspect of the van, but was inaccurate in delineating its ventral features. Using the shape prior, we  obtained an admissible result that is recognizable as a van. Although a van maintains a rigid structure, its shape in images can still be considered probabilistic due to uncertainty in pose.

We also processed pictures of leaves from the Caltech 101 image database\footnote{\url{http://www.vision.caltech.edu/Image_Datasets/leaves/}}. In the given examples, background objects interfere with the segmentation. Without a shape prior,  segmentations based on pixel intensities include these background distractors. To construct templates, we hand-traced six representative shapes for each of the two types of leaves. The regularized segmentations that we obtained were able to control for the background distractors.

\begin{figure}
\begin{center}
\includegraphics[width=0.3\textwidth]{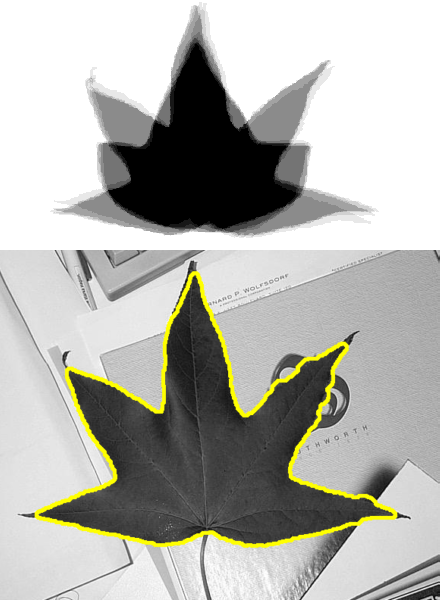}
\end{center}
\caption{\textbf{Segmentation using a hybrid shape-prior.} Using a combination of three and five-tip leaves, a ``hybrid" shape prior is constructed. Although the average of the shape templates is a leaf with seven tips, the kernel-density shape prior model is able to settle into a shape with the proper configuration of five tips.  }
\label{fig:fig4}
\end{figure}

In Fig~\ref{fig:fig4}, we segmented a five-pointed leaf using a prior composed of both three-tip and five-tip templates. Although the mean shape of the hybrid-shape prior is a leaf with seven tips, the algorithm relaxed into the correct five-tip state.

\section{Discussion}

We have presented a method for regularization of segmentation using a shape prior that is learned through kernel-density-estimation. Using a majorization-minimization trick, we showed how one can minimize the resulting energy, that includes a non-linear kernel-density term, using graph-cuts. Although there have been previous studies that include statistical shape priors in graph-cuts segmentation~\cite{zhu2007graph,cremers2006statistical},  these methods have used unimodal shape prior distributions where the input shape templates no longer retain their a-priori weights. Our shape-prior is a multi-modal kernel density estimator like the one used by~\citet{cremers2006kernel}, thereby enjoying many of the benefits outlined in ~\citet{cremers2006kernel}. 
 
   The advantage of our method however is speed, which can be a significant limitation in applications involving ``big data." It is also an issue in tracking applications, where many successive images need to be processed. Using an MM algorithm, we find surrogate energy functionals that can be minimized quickly using graph cuts. The surrogate energy functional~(Eq \ref{eqn:mmenergy}) contains an iteratively re-weighted shape regularization term, where shapes that are the most-similar the current segmentation estimate are the most influential. In our examples, convergence of segmentation defined as stationarity of the labeling typically takes place in few ($<10$) iterations.

Closer examination of Eq.~\ref{eqn:mmenergy} is revealing. Essentially, our algorithm takes advantage of the mathematical link between the relaxation of the nonlinear kernel-shape-prior energy functional of~\citet{cremers2006kernel} and iterative relaxation of the linear shape-prior energy functional found in many studies on incorporating shape priors into graph cuts~\cite{zhu2007graph,cremers2006statistical,vu2008shape}. As a result, our method has a computational complexity that is similar to that of these graph-cuts methods, which also rely on iterative template-alignment, while retaining many of the advantages of kernel-density estimation.

\subsection{Limitations}

As in other iterative schemes for image segmentation, our optimization method is local. Consequently, the iteration can become trapped in local minima, as in the level-set method of~\citet{cremers2006kernel}. It is important to note, however, that the notion of locality is different between this method and in the level-set method. In the level-set method, a boundary is propogated to steady state, where the boundary undergoes only small deformations between iterations.

 In this method, the boundary is able to jump between iteratations~\cite{chang2012tracking}. The locality of our approach refers to locality in the influence of the templates -- the weights -- as the segmentation iterates. Each iteration involves the optimization of an energy (Eq.~\ref{eqn:mmenergy}) that is linear in $U_\textrm{shape}$. While the method we provide for minimization of this energy is local, because of the locality of the template-alignment procedure, other non-local techniques for minimizing the surrogate energy~\cite{lempitsky2008image} may be used. In this manner, one could trade-off speed for globality.

To prevent poorly-aligned templates from dominating over the likelihood, particuarly in early iterations, one may treat $\beta$ as an annealment parameter. Please see \citet{vu2008shape}.

\subsection{Computational considerations}

Each iteration of the MM-algorithm is solved quickly using graph-cuts. Many efficient algorithms for this optimization exist, yielding excellent performance on modern hardware. There are many articles discussing the performance advantages of graph-cuts~\cite{boykov2001fast,grady2009piecewise}. For even more speed, there are GPU-accelerated approaches~\cite{vineet2008cuda}, and flow-reuse approaches~\cite{kohli2007dynamic}.

    The minimization of the surrogate energy is not the bottleneck of our method. To unleash the full potential of this shape regularization method, one needs to optimize the processing steps that are in the periphery the graph cut optimization -- the inference of the image intensity model, and the alignment of the templates. 
    
    As far as we know, other methods of shape-regularized segmentation also require these processing steps. Fortunately, these operations can be easily parallelized. Furthermore, they need not be repeated in every step of the algorithm. The image intensity parameters need only change if a significant proportion of the labels change. Similarly, the templates only need re-alignment if the  segmentation has significantly changed. To improve performance, one may calculate the transformation needed for a single template, and use that transformation for the entire set of templates. One may also use a fast moment-based method~\cite{jiang2008robust,pei1995image} to provide an initial state for shape alignment.

\subsection{Future directions}

The MM algorithm is useful for linearizing objective functionals. In this paper, we have presented it as a technique for linearizing a nonlinear shape prior, yielding a linearized surrogate energy functional. It may prove to be a useful tool for the graph-cut community for the relaxation other useful nonlinear functionals. Finally, the MM and EM algorithms may have use in inference of more-robust likelihood models. In our opinion, it is an avenue worth investigating.

\section{Acknowledgements}

JC acknowledges support by grant number T32GM008185 from
the National Institute of General Medical Sciences. JC and
TC also acknowledge support from the the National Science
Foundation through grants DMS-1032131 and DMS-1021818,
and from the Army Research Office through grant 58386MA.

\appendices

\section{Shape alignment}\label{sec:shapealignment}

\label{sec:newton}
In order to align the shape templates, we choose the affine transformation $\mathbb{T}(\sss)=\alpha\mathbf{R}(\boldsymbol\omega)\left(\sss-\ccc\right)$ that minimizes the shape energy
 \begin{align*}
\lefteqn{U_{\textrm{shape}}\left(\mathbb{T}\left(\ccc,\boldsymbol\omega,\alpha\right)\right)=}\\
&\qquad\overbrace{\int_{\mathbb{R}^d} \Big(\chi_\Omega(\sss)-\chi_\Lambda(\mathbb{T}(\sss))\Big)^2\Big|\phi_\Lambda (\mathbb{T}(\sss))\Big|^\lambda d\sss}^{\textrm{mass mismatch}} \\
&\qquad+ \overbrace{\int_{\mathbb{R}^d}\delta\left(\phi_\Omega(\sss)\right) \Big|\phi_{\Lambda}(\mathbb{T}(\sss))\Big|^\lambda d\sss}^{\textrm{boundary mismatch}}.
\end{align*}
In this expression, we have rewritten the contour integral on $\partial\Omega$ in terms of an integral over the zero-level set of $\phi_\Omega$.
Here, we develop a local Newton-Raphson algorithm for finding the optimal transform. 
Let us denote the $2\cdot d$ column vector of transformation parameters $\boldsymbol\varphi=\left[\alpha\ \ \mathbf{c}\ \ \ \boldsymbol\omega \right] $. We estimate $\boldsymbol\varphi$ using the iterative updates
\[ \boldsymbol\varphi_{n+1} = \boldsymbol\varphi_n - \left[\mathbf{H}_{\boldsymbol\varphi}\left(\boldsymbol\varphi_n \right)\right]^{-1}\nabla_{\boldsymbol\varphi} U_{\textrm{shape}}(\boldsymbol\varphi_n)  \]
where
\[ \nabla_{\boldsymbol\varphi}E = \left[ \frac{\partial}{\partial\alpha}\ \ \nabla_{\mathbf{c}}^\textrm{T} \ \ \nabla_{\boldsymbol\omega}^\textrm{T} \right]^\textrm{T}U_{\textrm{shape}} \]
and
\[
\mathbf{H}_{\boldsymbol\varphi} = \left[ 
\begin{matrix}
 \displaystyle\frac{\partial^2}{\partial \alpha^2}   & \displaystyle\nabla_{\mathbf{c}}^\textrm{T}\frac{\partial}{\partial\alpha}& \displaystyle\nabla_{\boldsymbol\omega}^\textrm{T}\frac{\partial}{\partial\alpha} \\ 
 \displaystyle\nabla_{\mathbf{c}}\frac{\partial}{\partial\alpha} &  \displaystyle\nabla_{\mathbf{c}}\nabla_{\mathbf{c}}^\textrm{T} & \nabla_{\boldsymbol\omega}^\textrm{T}\nabla_{\mathbf{c}}\\
  \displaystyle\nabla_{\boldsymbol\omega}\frac{\partial}{\partial\alpha} & \nabla_{\boldsymbol\omega}\nabla_{\mathbf{c}}^\textrm{T}& \displaystyle\nabla_{\boldsymbol\omega}\nabla_{\boldsymbol\omega}^\textrm{T}
 \end{matrix} \right] U_{\textrm{shape}}.
  \]
To populate the matrix, we need to calculate the gradients with respect to $\alpha$, $\mathbf{c}$, and $\boldsymbol\omega$. For $\lambda>0$, the first-order derivatives take the form 
\begin{align*} 
\nabla_{\boldsymbol\varphi_i}U_{\textrm{shape}} &=  \lambda\int_{\mathbb{R}^d} \left[ \Big(\chi_\Omega(\sss)-\chi_\Lambda(\sssp)\Big)^2+ \delta\big(\phi_\Omega(\sss)\big) \right] \\
&\qquad \times\Big|\phi_{\Lambda}(\sssp)\Big|^{\lambda-1} \sgn(\phi_\Lambda(\sssp))\underbrace{\frac{\partial\mathbb{T}}{\partial\boldsymbol\varphi_i}}_{p\times d}\underbrace{ \nabla \phi_{\Lambda}(\sssp)}_{d\times 1}d\sss
\end{align*}
The sign function comes about from differentiation of the absolute value function in the distributional sense. 

For $\lambda\neq1$, the second-order derivatives constitute tensors that take the form
\begin{align*}
\lefteqn{\nabla_{\boldsymbol\varphi_j}^\textrm{T}\nabla_{\boldsymbol\varphi_i}U_{\textrm{shape}}  =}\\
&\qquad \lambda(\lambda-1)\int_{\mathbb{R}^d} \left[ \left(\chi_\Omega(\sss) - \chi_\Lambda(\sssp) \right)^2+\delta\left(\phi_\Omega(\sss) \right)\right] \\
&\qquad\times \left| \phi_\Lambda(\sssp)\right|^{\lambda-2} \frac{\partial\mathbb{T}}{\partial\boldsymbol\varphi_j} \nabla\phi_\Lambda(\sssp)\nabla^\mathbf{T}\phi_\Lambda(\sssp)\left(\frac{\partial\mathbb{T}}{\partial\boldsymbol\varphi_i}  \right)^\mathbf{T}d\sss \\
&+2\lambda\int_{\mathbb{R}^d}\delta\left(\phi_\Lambda(\sssp) \right) \left[ \Big(\chi_\Omega(\sss)-\chi_\Lambda(\sssp)\Big)^2+ \delta\big(\phi_\Omega(\sss)\big) \right] \\
&\qquad\times\left| \phi_\Lambda(\sssp)\right|^{\lambda-1}\frac{\partial\mathbb{T}}{\partial\boldsymbol\varphi_i} \nabla \phi_{\Lambda}(\sssp)\nabla^{\mathbf{T}}\phi_\Lambda(\sssp)\left( \frac{\partial\mathbb{T}}{\partial\boldsymbol\varphi_j} \right)^\textrm{T}  d\sss\\
&+\lambda\int_{\mathbb{R}^d} \left[ \Big(\chi_\Omega(\sss)-\chi_\Lambda(\sssp)\Big)^2+ \delta\big(\phi_\Omega(\sss)\big) \right]\left| \phi_\Lambda(\sssp)\right|^{\lambda-1}\\
&\qquad\times\sgn\phi_\Lambda(\sssp) \frac{\partial\mathbb{T}}{\partial\boldsymbol\varphi_i}\nabla \nabla\phi_\Lambda(\sssp) \left(\frac{\partial\mathbb{T}}{\partial\boldsymbol\varphi_j}\right)^\textrm{T}d\sss\\
&+\lambda\int_{\mathbb{R}^d} \left[ \Big(\chi_\Omega(\sss)-\chi_\Lambda(\sssp)\Big)^2+ \delta\big(\phi_\Omega(\sss)\big) \right]\\
&\qquad\times\left| \phi_\Lambda(\sssp)\right|^{\lambda-1} \sgn\phi_\Lambda(\sssp)\overbrace{\frac{\partial^2\mathbb{T}}{\partial\boldsymbol\varphi^\textrm{T}_j\partial\boldsymbol\varphi_i}}^{p\times p\times d}: \nabla \phi_{\Lambda}(\sssp) d\sss.\\
\end{align*}
If $\lambda=1$, the first term is zero and the following term is added
\begin{align*}
&-2\lambda\int_{\mathbb{R}^d}\delta\left(\phi_\Lambda(\sssp)\right)\Big(\chi_\Omega(\sss)-\chi_\Lambda(\sssp) \Big)\sgn(\phi_\Lambda(\sssp))\\
&\qquad\qquad{\frac{\partial\mathbb{T}}{\partial\boldsymbol\varphi_j}}{ \nabla \phi_{\Lambda}(\sssp)}\left({\frac{\partial\mathbb{T}}{\partial\boldsymbol\varphi_i}}{ \nabla \phi_{\Lambda}(\sssp)}\right)^{\mathbf{T}}d\sss.
\end{align*}
The contour integrals in these expressions can be calculated using a regularized version of the Dirac delta function such as
\[ \delta_\epsilon(x) = \frac{1}{\pi}\frac{\epsilon}{\epsilon^2+x^2} \qquad\epsilon\to0^+. \]
Similarly, the characteristic function can be interpreted as the Heaviside function acting on the signed-distance shape embedding, which can be approximated using an approximation of the Heaviside function such as
\[ H_\epsilon(x) = \frac{1}{2} + \frac{1}{\pi}\arctan\frac{x}{\epsilon} \qquad \epsilon\to0^+. \]

The nonzero transformation derivatives are as follows
\begin{align*}\frac{\partial\mathbb{T}}{\partial\mathbf{c}} &= -\alpha \mathbf{R} \\
\frac{\partial\mathbb{T}}{\partial\alpha} &=\mathbf{R}(\sss-\ccc) \\
\frac{\partial\mathbb{T}}{\partial\boldsymbol\omega}&=\alpha\frac{\partial\mathbf{R}}{\partial\boldsymbol\omega}(\sss-\ccc) \\
\frac{\partial^2\mathbb{T}}{\partial\boldsymbol\omega\partial\mathbf{c}} &= -\alpha \frac{\partial\mathbf{R}}{\partial\boldsymbol\omega} \\
\frac{\partial^2\mathbb{T}}{\partial\alpha\partial\boldsymbol\omega} &=\frac{\partial\mathbf{R}}{\partial\boldsymbol\omega}(\sss-\ccc) \\
\frac{\partial^2\mathbb{T}}{\partial\boldsymbol\omega\partial\boldsymbol\omega}&=\alpha\frac{\partial^2\mathbf{R}}{\partial\boldsymbol\omega\partial\boldsymbol\omega}(\sss-\ccc) \\
\frac{\partial^2\mathbb{T}}{\partial\alpha\partial\mathbf{c}}&=-\mathbf{R}.
\end{align*}

\begin{figure}
\begin{center}
\includegraphics[width=21pc]{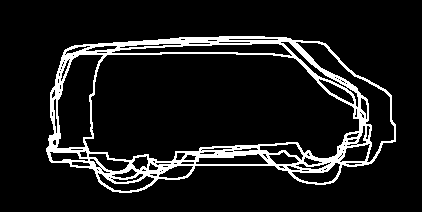}
\caption{\textbf{Inputted shape templates for a van.} These templates are aligned using the method detailed in appendix~\ref{sec:newton} in order to generate the aligned templates of Fig.~\ref{fig:fig3}.}
\label{fig:fig5}
\end{center}
\end{figure}

The results of this alignment method are shown in the third column Fig.~\ref{fig:fig3}. As an example, the raw input templates for the vans in this figure are shown in Fig.~\ref{fig:fig5}.


\section{MM algorithm for iterative graph cuts}\label{sec:graphembed}

The shape term $\log\sum w_j K(\Omega,\Omega_j)$ can make
minimization of the energy difficult, since its formulation involves a
sum within a logarithm, making the energy functional nonlinear
with respect to the labeling of pixels (into background and foreground) in the image.

 To linearize the shape contribution, we will derive a surrogate function with
separated terms. A function $f(x|x_k)$, with fixed $x_k$, is said to majorize a function
$g(x)$ at $x_k$  if the following holds~\citep{hunter2004tutorial}
\begin{align}
g(x)&\leq f(x|x_k) \label{eqn:ineq}  \\
f(x_k)&=g(x_k|x_k) \label{eqn:eq}.
\end{align}
We wish to perform iterative inference by finding a sequence of segmentations $\Omega^{(n+1)} = \arg\min_{\Omega}
Q(\Omega|\Omega^{(n)})$, where $Q(\Omega|\Omega^{(n)})$ majorizes
Eq.~\ref{eqn:energysum}. By the descent property of the MM algorithm~\citep{hunter2004tutorial}, this sequence converges to a local minimum.

For any convex function $f(x)$, the following holds~\citep{hunter2004tutorial}
\[
f\left(\sum_i \alpha_it_i\right)\leq \sum_i\alpha_i f(t_i).
\]
Noting that $-\log(\cdot)$ is convex, we have
\begin{align}
\lefteqn{-\log\overbrace{\sum_{j=1}^J w_j K(\Omega,\Omega_j)}^{\text{shape kernel density}}\leq}\nonumber\\ & -\sum_{j=1}^J \frac{w_j K(\Omega^{(n)},\Omega_j)}{\sum_{k=1}^J w_kK(\Omega^{(n)},\Omega_k)}\nonumber\\
&\qquad\times\log\left[ \frac{\sum_{k=1}^J w_kK(\Omega^{(n)},\Omega_k)}{w_j K(\Omega^{(n)},\Omega_j)} w_j K(\Omega,\Omega_j) \right]\nonumber 
\end{align}
\begin{align}
&= - \sum_{j=1}^J \frac{w_j K(\Omega^{(n)},\Omega_j)}{\sum_{k=1}^J w_kK(\Omega^{(n)},\Omega_k)}\log K(\Omega,\Omega_j) \nonumber\\&\qquad+ \text{const} \nonumber\\
&= \underbrace{\frac{\beta}{2} \sum _{j=1}^J \frac{w_j K(\Omega^{(n)},\Omega_j)}{\sum_{k=1}^J w_kK(\Omega^{(n)},\Omega_k)}U_{\textrm{shape}}(\Omega,\Omega_j)}_{\text{separated log shape kernel density}}\nonumber\\
&\qquad+ \text{const}\label{eqn:majorizor},
\end{align}verifying that the inequality condition~(Eq.~\ref{eqn:ineq}) holds.
In the case that $\Omega=\Omega^{(n)}$, we have
\begin{align*}
\lefteqn{ -\sum_{j=1}^J \frac{w_j K(\Omega^{(n)},\Omega_j)}{\sum_{k=1}^J w_kK(\Omega^{(n)},\Omega_k)}}\nonumber\\
&\qquad\times\log\left[ \frac{\sum_{k=1}^J w_kK(\Omega^{(n)},\Omega_k)}{w_j K(\Omega^{(n)},\Omega_j)} w_j K(\Omega^{(n)},\Omega_j) \right]\nonumber \\
&= -\log\sum_{j=1}^J w_j K(\Omega^{(n)},\Omega_j),
\end{align*}
verifying that the equality condition~(Eq.~\ref{eqn:eq}) holds.
So the two majorizing conditions~(\ref{eqn:ineq},\ref{eqn:eq}) are met, and we can minimize our original energy by iteratively minimizing 
\begin{align}
\lefteqn{Q(\Omega|\Omega^{(n)})=-\sum_{\sss\in\Omega}\log p_\Omega(I(\sss)|\boldsymbol\theta)-\sum_{\sss\in\Delta}\log p_\Delta(I(\sss)|\boldsymbol\theta)}\nonumber\\
&\quad-\log p(\boldsymbol\theta|\Omega) \nonumber\\
 &\quad+ \frac{\beta}{2} \sum _{j=1}^J \frac{w_j K(\Omega^{(n)},\Omega_j)}{\sum_{k=1}^J w_kK(\Omega^{(n)},\Omega_k)}U_{\textrm{shape}}(\Omega,\Omega_j).
\label{eqn:mmenergy2}
\end{align}
Because the distance function can be written as a sum over  vertices and edges of a graph, so can Eq.~\ref{eqn:mmenergy2}. As a result, it is possible to
minimize Eq.~\ref{eqn:mmenergy2} within the graph cuts framework.

\bibliography{segmentation}

\end{document}